\documentclass[runningheads]{llncs}
\usepackage{graphicx}
\usepackage{times}
\usepackage{latexsym}
\usepackage{multirow}
\usepackage{array}
\usepackage[normalem]{ulem}
\useunder{\uline}{\ul}{}
\usepackage{booktabs}
\usepackage{amsmath}
\usepackage{orcidlink}

\usepackage{subcaption}
\usepackage{graphicx}

\usepackage{xcolor}

\usepackage[T1]{fontenc}
\usepackage[utf8]{inputenc}

\usepackage{microtype}
\usepackage[misc]{ifsym}

\usepackage[draft,textsize=footnotesize,textwidth=15mm]{todonotes}
\begin{document}
%
%PEACE: Using Causal Cues for Cross-Platform Hate Speech Detection \\
\title{PEACE: Cross-Platform Hate Speech Detection \\ - A Causality-guided Framework}

% %
% %\titlerunning{Abbreviated paper title}
% % If the paper title is too long for the running head, you can set
% % an abbreviated paper title here
\titlerunning{PEACE}
\authorrunning{Sheth and Kumarage et al.}
\author{Paras Sheth\tocauthor{Paras~Sheth}\thanks{Both authors contributed equally.}\Letter\inst{1}\orcidlink{0000-0002-6186-6946} \and
Tharindu Kumarage\tocauthor{Tharindu~Kumarage}$^\star$\inst{1}\orcidlink{0000-0002-9148-0710} \and
Raha Moraffah\tocauthor{Raha~Moraffah}\inst{1}\orcidlink{0000-0002-6891-2925} \and
Aman Chadha\tocauthor{Aman~Chadha}\thanks{Work does not relate to the position at Amazon}\inst{2,3}\orcidlink{0000-0001-6621-9003} \and
Huan Liu\tocauthor{Huan~Liu}\inst{1}\orcidlink{0000-0002-3264-7904}}

\institute{Arizona State University, Tempe, AZ, USA \and
Stanford University, Stanford, CA, USA \and Amazon Alexa AI, Sunnyvale, CA, USA  \\
\email{\{psheth5,kskumara,rmoraffa,huanliu\}@asu.edu}\\
\email{hi@aman.ai}}

\toctitle{PEACE: Cross-Platform Hate Speech Detection \\ - A Causality-guided Framework}
\maketitle              % typeset the header of the contribution
\begin{abstract}
Hate speech detection refers to the task of detecting hateful content that aims at denigrating an individual or a group based on their religion, gender, sexual orientation, or other characteristics. Due to the different policies of the platforms, different groups of people express hate in different ways. Furthermore, due to the lack of labeled data in some platforms it becomes challenging to build hate speech detection models. To this end, we revisit if we can learn a generalizable hate speech detection model for the cross platform setting, where we train the model on the data from one (source) platform and generalize the model across multiple (target) platforms. 
%% HL As a result, an automated hate speech detection model trained on the data for one platform \HL{may} fail to generalize across different platforms. These models fail to generalize because 
Existing generalization models rely on linguistic cues or auxiliary information, making them biased towards certain tags or certain kinds of words (e.g., abusive words) on the source platform and thus not applicable to the target platforms.
%% HL in the presence of distribution shifts that may arise from the way different users express themselves across platforms. 
Inspired by social and psychological theories, we endeavor to explore if there exist inherent causal cues that can be leveraged to learn generalizable representations for detecting hate speech across these distribution shifts. To this end, we propose a causality-guided framework, \textbf{PEACE}, that identifies and leverages two intrinsic causal cues omnipresent in hateful content: the overall sentiment and the aggression in the text. We conduct extensive experiments across multiple platforms (representing the distribution shift) showing if causal cues can help cross-platform generalization.
\end{abstract}
\section{Introduction}
\textbf{Warning:} \textit{this paper contains contents that may be offensive or upsetting.}

Social media sites have served as global platforms for users to express and freely share their opinions. However, some people utilize these platforms to share hateful content targeted toward other individuals or groups based on their religion, gender, or other characteristics resulting in the generation and spread of hate speech. Failing to moderate online hate speech has shown to have negative impacts in real world scenarios, ranging from mass lynchings to global increase in violence toward minorities~\cite{laub2019hate}. Thus, building hate speech detection models has become a necessity to limit the spread of hatred. Recent years have witnessed the development of these models across disciplines~\cite{paz2020hate,fortuna2018survey,williams2020hate,alkomah2022literature}.

% Considerate efforts have been made to develop hate speech detection algorithms. 
Hate speech varies based on the platform and the specific targets of the speech, influenced by factors such as social norms, cultural practices, and legal frameworks. Platforms with strict regulation policies may lead to users expressing hate in subtle ways (e.g., sarcasm), while platforms with lenient policies may have more explicit language. Collecting large labeled datasets for hate speech detection models is challenging due to the emotional burden of labeling and the requirement for skilled annotators~\cite{macavaney2019hate}. One solution is to train a generalizable model under a cross-platform setting, leveraging the labeled data from other platforms.
% However, due to social, cultural, and legal contexts, hate speech can vary by targets of hate, and platform. For instance, in platforms with strict regulation policies, users resort to subtle ways to express hate (e.g., using sarcasm), whereas, in platforms with lenient policies, users are more explicit and utilize profane words. Moreover, different platforms may focus on different targets of hate. When developing hate speech detection model for a platform one problem that arises is the limited availability of labeled data. Labeling data as hate speech can be emotionally taxing and may require significant resources, including human annotators with expertise in identifying hate speech making it challenging to collect a large scale dataset of hate speech. One way to circumvent this problem is to train a generalizable hate speech detection model by leveraging data from multiple platforms.

% Furthermore, lack of data hinders the ability to develop and deploy hate speech detection models across platforms. One way to circumvent this problem is to  

% For instance,  on social media platforms like Facebook, hate speech is mainly directed toward a particular group such as muslims~\cite{ghasiya2022rapid,karjo2020hate}. In contrast, on gaming platforms such as YouTube Gaming, hate speech is often directed toward players based on race, gender, or sexual orientation~\cite{chatzakou2017measuring,doring2020gendered}. Furthermore,

%  Thus, a generalizable model should perform well across different platforms.

Recent works developed to improve the cross-platform performance utilize either linguistic cues such as vocabulary~\cite{ramponi-tonelli-2022-features} or Parts-Of-Speech (POS) tags~\cite{markov2021exploring}. Another direction leverages datasets with auxiliary information such as implications of various hate posts~\cite{kim2022generalizable} or the groups or individuals attacked in the hate post~\cite{kennedy2018gab}. Although effective, these methods suffer from shortcomings, such as linguistic methods form spurious correlations towards certain POS tags (e.g., adjectives and adverbs) or a particular category of words (e.g., abusive words). In addition, methods that utilize auxiliary information (e.g., implications of the post or the target(s)) are not extendable as the auxiliary information may not be available for large datasets or different platforms.

In contrast to previous approaches, we contend that identifying inherent causal cues is necessary for developing effective cross-platform hate speech detection models that can distinguish between hateful and non-hateful content. Since causal cues are immune to distribution shifts~\cite{buhlmann2020invariance}, leveraging them for learning the representations can aid in better generalization. Various studies in social sciences and psychology verify the existence of several cues that can aid in detecting hate~\cite{sengupta2022does,craig2002examining,krahe2020social,bauwelinck2019measuring,zhou2021hate} such as the hater's prior history, the conversational thread, overall sentiment, and aggression in the text. However, when dealing with a cross-platform setting, several cues may not be accessible. For instance, not all platforms allow access to user history or the entire conversation thread. Thus, we propose to leverage two causal cues namely, the overall sentiment and the aggression in the text. Both these cues can be measured easily with the aid of aggression detection tasks~\cite{aroyehun2018aggression} and sentiment analysis task~\cite{yue2019survey}. Moreover, both aggression and sentiment are tightly linked to hate speech. For instance, due to the anonymity on online platforms, users adopt more aggressive behavior when expressing hatred towards someone~\cite{rosner2016verbal}. Thus, the aggression in the content could act as a causal cue to indicate hate. Similarly, hateful content is meant to denigrate someone. Thus, the sentiment also serves as a causal cue~\cite{rodriguez2019automatic}.

% The use of causal cues can be beneficial in terms of generalization due to their invariant nature~\cite{buhlmann2020invariance}, i.e., these cues are reasonably invariant to distribution shifts. Thus, they can be leveraged to enhance the generalization performance. For instance, in the context of hate speech, due to the anonymity on online platforms, users adopt more aggressive behavior when expressing hatred towards someone. Thus, the aggression present in the content could act as a causal cue to indicate hate. Similarly, hateful content is designed with the intention to denigrate someone. Thus, the overall sentiment of the content can serve as another causal cue. Our hypothesis is further corroborated by various social science and psychology theories that verify that aggression and sentiment are inherent to hate speech and can serve as causal cues~\cite{}.

To this end, we propose a novel causality-guided framework, namely, \textbf{P}latform-ind\textbf{E}pendent c\textbf{A}usal \textbf{C}ues for generalizable hat\textbf{E} speech detection \textbf{PEACE}\footnote{The code for PEACE can be accessed from: \url{https://github.com/paras2612/PEACE}}, that leverages the overall sentiment and the aggression in the text, to learn generalizable representations for hate speech detection across different platforms. We summarize our main contributions as follows:
\begin{itemize}
    \item We identify two causal cues, namely, the overall sentiment and the aggression in the text content, to learn generalizable representations for hate speech detection. 
    \item We propose a novel framework, namely, \textbf{PEACE} consisting of multiple modules to capture the essential latent features helpful for predicting sentiment and aggression. Finally, we utilize these features and the original content to learn generalizable representations for hate speech detection.
    \item Experimental results on five different platforms demonstrate that \textbf{PEACE} achieves state-of-the-art performance compared with vital baselines, and further experiments highlight the importance of each causal cue and interpretability of \textbf{PEACE}.
\end{itemize}

\section{Related Work}
Social media provides a vast and diverse medium for users to interact with each other effectively and share their opinions. Unfortunately, however, a large share of users exploits these platforms to spread and share hateful content mainly directed toward an individual or a group of people. Considering the massive volume of online posts, it is impractical to moderate them manually. To address this shortcoming, researchers have proposed various methods ranging from lexical-based approaches~\cite{gitari2015lexicon,markov2021exploring,wiegand2018inducing} to deep learning-based approaches~\cite{mazari2023bert,del2023socialhaterbert,roy2021leveraging}.

However, these models have been shown to possess poor generalization capabilities. Hate speech on social media is highly volatile and is constantly evolving. A hate speech detection model that fails to generalize well may exhibit poor detection skills when dealing with a new topic of hate~\cite{pamungkas2021joint,del2017hate} or when dealing with different styles of expressing hate~\cite{corazza2019cross,ali2022hate}, thus making it critical to develop generalizable hate speech detection models. Over recent years there has been an increase in developing generalizable models.

Generalizable hate speech detection methods can be broadly classified into two parts, namely models that leverage auxiliary information such as implications of hate posts~\cite{kim2022generalizable}, information of the dataset annotators~\cite{yin2022annobert}, or user attributes~\cite{del2023socialhaterbert}. For instance, the authors of the work~\cite{kim2022generalizable} proposed a generalizable model for implicit hate speech detection that utilizes the implications of hateful posts and learns contrastive pairs for a more generalizable representation of the hate content. Similarly, the authors of the work~\cite{yin2022annobert} argue that when dealing with subjective tasks such as hate speech detection, it is hard to achieve agreement amongst annotators. To this end, they propose leveraging the annotator's characteristics and the ground truth label during the training to learn better representations and improve hate speech detection. Unlike annotators' information, the authors of~\cite{del2023socialhaterbert} trained a bert model with users' profiles and related social environment and generated tweets to infer better representations for hate speech detection. Although these models have improved generalizability, the auxiliary information utilized may not be easily accessible and challenging to get when dealing with cross-platform settings.

Since language models are trained on large corpora, they exhibit some generalization prowess~\cite{tamkin2020investigating}. However, the generalization can be improved by finetuning these models on datasets related to a specific downstream task. Thus, the second category leverages language models such as BERT~\cite{devlin2018bert} and finetuning them on large hate speech corpora~\cite{caselli2020hatebert,mathew2021hatexplain}. For instance, the authors of~\cite{caselli2020hatebert} finetuned a BERT model on approximately 1.6 million hateful data points from Reddit and generated HateBERT, a state-of-the-art model for hate speech detection. Similarly, the authors of~\cite{mathew2021hatexplain} finetuned BERT for explainable hate speech detection. Aside from these works, some methods focus on leveraging lexical cues such as vocabulary used~\cite{schmidt2017survey}, emotion words, and different POS tags in the content~\cite{markov2021exploring}, the target-specific keyphrases~\cite{elsherief2018hate}.

Although these methods have been shown to improve hate speech detection capabilities, these require large labeled corpora for finetuning language models, which may not be feasible in the real-world setting as the number of posts generated in a moment is extremely large or rely on lexical features which may not aid as a lot of the social media posts are filled with grammatical inconsistencies (such as misspelled words). In this work, inspired by works in social and psychological fields, we leverage inherent characteristics readily available in the text to learn generalizable representations, such as the aggression and the overall sentiment of the text.

\section{Methodology}
% \raha{follow this for example: 1. main framework 1.1.cue extraction module 1.1.1 cue1 1.1.2. cue2 1.1.3 integration 1.2 classifier}

% \subsection{Main Framework}

\begin{figure}[h]
     \centering
     \includegraphics[width=0.75\textwidth]{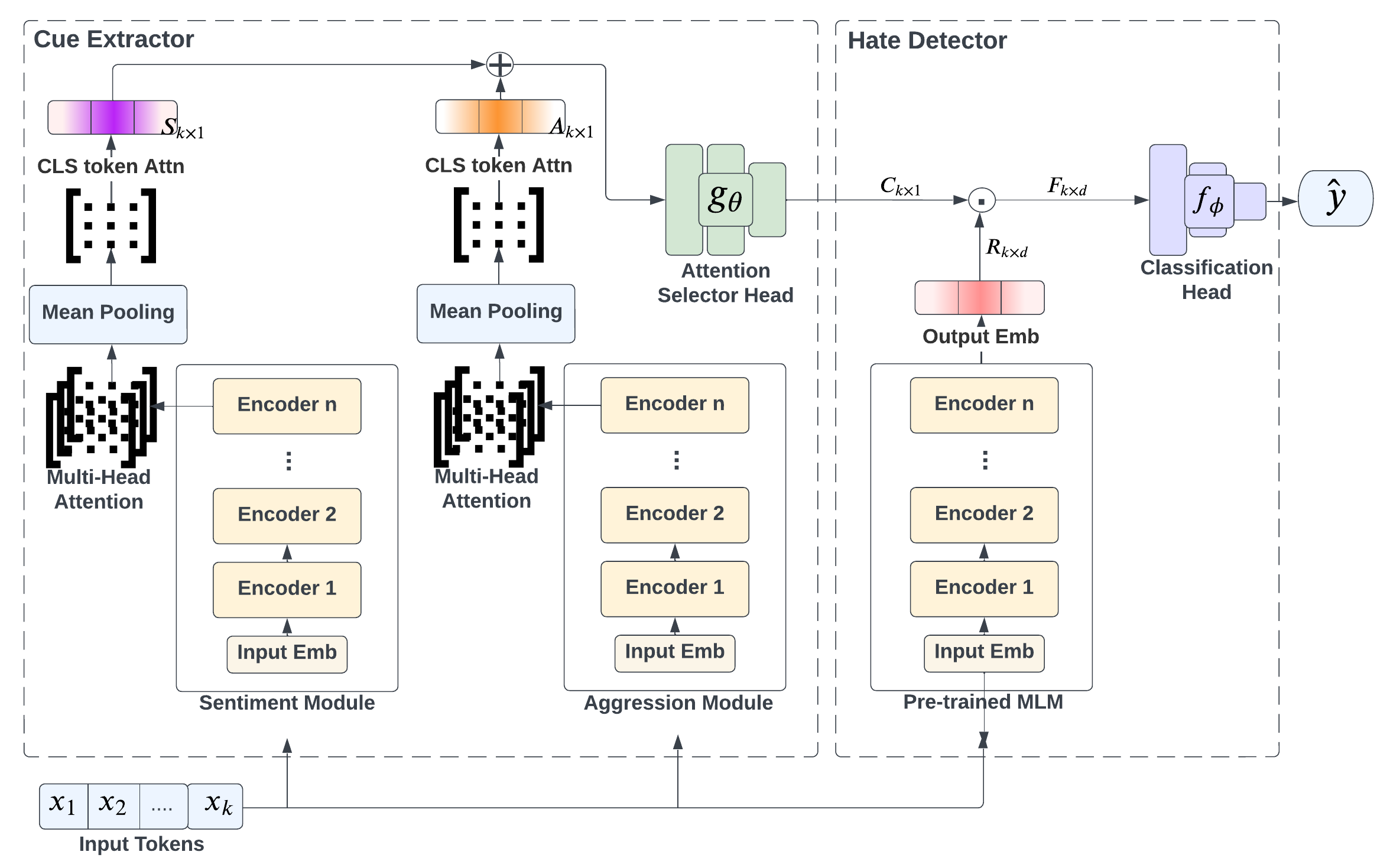}
     \caption{Proposed framework architecture for \textbf{PEACE}. The pre-trained sentiment and aggression modules guide the representation learning process to ensure generalizability.}
\label{fig:modelarchi}
\end{figure}

This section describes the methodology behind our \textbf{PEACE} framework. As shown in Figure \ref{fig:modelarchi} the framework consists of two major components: (i) a cue extractor component and (ii) a hate detector component. The cue extractor component extracts the proposed innate cues, sentiment, and aggression. Moreover, this component is responsible for navigating the hate detector component toward learning a cross-platform generalized representation for hate speech detection. Consequently, the hate detector component classifies a given input to hate or non-hate classes while attending to the causal guidance of the cue extractor. In the subsequent sections, we discuss the cue extractor and hate detector components in detail.  

\subsection{Causal Cue Extraction}
We propose utilizing sentiment and aggression as two inherent causal cues for learning generalizable representations for better hate speech detection. Therefore, the cue extractor consists of two modules, one for extracting sentiment and one for aggression. Given an input text $X = (x_{1}, x_{2}, ..., x_{k})$,  the purpose of the cue extractor model is to generate an attention vector $C_{k\times1}$ where $k$ is the input sequence length. And here, the vector $C_{k\times1}$ should represent an accumulation of sentiment and aggression score for each token in the sequence $X$, i.e., for a given token in the input $X$, $C_{k\times1}$ contains how vital that token is towards the overall input's sentiment and/or aggression. We will first discuss the architecture of each cue module (sentiment and aggression) and then elaborate on how the attention vector $C_{k\times1}$ is generated. 

\subsubsection{Sentiment Module}
The sentiment module is a transformer encoder stack with $n$ encoders that have learned a function $s_{\gamma}$ such that given an input text $X = (x_{1}, x_{2}, ..., x_{k})$, it can classify the sentiment of $X$, i.e., this module is a pre-trained transformer-based large language model finetuned for the sentiment detection downstream task where given an input text $X$, it predicts the sentiment label $y$ (positive, neutral, negative), $y = s_{\gamma}(X)$. 

\subsubsection{Aggression Module}

Similarly, the aggression module is also a transformer encoder stack with $n$ encoders that have learned a function $a_{\lambda}$ such that given an input text $X = (x_{1}, x_{2}, ..., x_{k})$, it can classify whether $X$ contains aggressive speech, i.e., this module is a pre-trained transformer-based large language model finetuned for the aggression detection downstream task where given an input text $X$, it predicts the aggression label $y$ (aggressive, non-aggressive), $y = a_{\lambda}(X)$. 

 And it is essential to note here that the cue extraction module's wights are frozen when we conduct the end-to-end training of the hate detector component, i.e., we don't finetune the sentiment and aggression modules with the hate speech data. 

% It is important to note here that the proposed methodology is to extract and utilize causal cues as an attention vector of size $k$, where $k$ is the size of the original input sequence $(x_{1}, x_{2}, ..., x_{k})$. Therefore, we only considered the finetuned language model block in both sentiment and aggression modules, not the classification head used for finetuning the corresponding task. 

\subsubsection{Attention Extraction for Individual Causal Cues}

As mentioned above, the cue extractor component aims to integrate the two cue modules, sentiment, and aggression, towards generating the final causal cue guidance as an attention vector $C_{k\times1}$. The first step towards this objective is extracting each individual attention vector from the cue modules. Since both the sentiment and aggression cue modules are same-sized transformer encoder stacks ($n$-encoders), the attention extraction process is the same for both modules. Let's take the sentiment cue module; it contains $n$-encoder blocks and thus consists of $n$ multi-head attention layers. The multi-head attention layer of a given encoder block can be defined as the Equation \ref{eq:multihead}.

\begin{equation}
\label{eq:multihead}
  \begin{aligned}
   MultiHead(Q,K,V) = head_1(Q,K,V) 
    \oplus ... head_n(Q,K,V)\\ 
    where; \quad head_i(Q,K,V) = softmax(\frac{QK^T}{\sqrt{d_i}})V
  \end{aligned}  
\end{equation}

Here $Q, K, V$ are Query, Key, and Value vectors of the transformer block $i$, and $d_i$ is the hidden state size \cite{vaswani2017attention}. 

Our goal in using the sentiment cue module attention is to figure out the words/phrases in the input text that has particular importance towards the sentiment of the text. Therefore, we need to consider an encoder block that gives comprehensive attention to the whole input. Previous research shows that the attention heads in the BERT model's last encoder block have very broad attention - i.e., attending broadly to the entire input \cite{clark2019does}. The architecture we consider for the sentiment module is similar to the BERT architecture (transformer encoder blocks); thus, we select the last ($n^{th}$) encoder block's multi-head attention layer as the candidate to extract the final attention from the sentiment module. We take the mean pooling output of the $n^{th}$ block's multi-headed attention layer as a matrix $M_{k\times k}$ where $k$ is the input sequence length. 

\begin{equation}
	M_{k\times k} = Mean(MultiHead_n(Q,K,V))
\end{equation}

Then the final attention vector $S_{k\times1}$ for the input sequence is taken by selecting the attention at CLS token of the matrix $M_{k\times k}$. Following the same process, we extract the aggression attention vector $A_{k\times1}$ from the aggression cue module.

\subsubsection{Cue Integration}

The final step towards creating the attention vector $C_{k\times1}$ is to aggregate each attention vector we get from cue modules. i.e., we need to weigh and aggregate the token attentions from each cue module to get the final accumulated attention vector $C_{k\times1}$. Once the representative attention vectors from both sentiment and aggression modules are extracted, we input the concatenated vectors through the attention selector head ($g_{\theta}$). The attention selector head is a fully connected neural network that takes concatenated aggression and sentiment attention to map the final attention vector $C_{k\times1}$. 

\begin{equation}
	C_{k\times1} = g_{\theta}([S_{k\times1} \oplus A_{k\times1}])
\end{equation}

The intuition behind the attention selector head is that we need our framework to learn how to weigh the sentiment and aggression cues relevant to the context of the given input. For example, there can be cases where aggression could be the stronger cue towards hate speech than sentiment or vice versa. 
% \acil{Offer an example for the above senario.}

\subsection{Hate Detector}

The hate detector component consists of a similar transformer encoder stack to learn the semantic representation of the given input. However, the output of the cue detector component, attention vector $C_{k\times1}$, will be provided as an auxiliary signal. We select the representation learned by the hate detector blocks as $R_{k\times d}$ where $k$ is the sequence length, and $d$ is the hidden state size of an encoder block. Then the extracted attention is used to navigate the hate detector to adjust the representation to incorporate the causal cues. The final representation $F_{k\times d}$ is calculated as;  $F_{k\times d} = R_{k\times d} \odot C_{k\times1}$. 
% \acil{Would be great to label $F_{k\times d}, R_{k\times d}, C_{k\times1}$ in fig 1.}
% \acil{Are you doing point-wise multiplication of R and C? If so, mention that explicitly or use the appropriate symbol (as in fig 1).}
Then the representation corresponding to the end of the sequence token ($F_{1\times d}^{CLS}$) is passed through the classification head ($f_{\phi}$). The classification head ($f_{\phi}$) is a fully connected neural network that takes the learned semantic embedding as the input and predicts the hate label $\hat{y}$ as $\hat{y} = f_{\phi}(F_{1\times d}^{CLS})$.

The overall framework is trained via the cross-entropy loss for the classification, where $y$ is the ground truth. 

\begin{equation}
	L = -\sum_{i}y_{i}\log(\hat{y}_{i})
\end{equation}

% \raha{please follow a model paper, this section should be called Experiments not Experiment settings}
\section{Experiments}
This section discusses the experimental settings used to validate our framework, including the datasets and evaluation metrics used, and the baselines, followed by a detailed analysis of the experiments. We conducted a series of experiments to understand whether the identified causal cues, namely the sentiment and the aggression in the text, can aid in learning generalizable representations for hate speech detection and answer the following research questions.
% \acil{Rephrase the above line.}
\begin{itemize}
	\item \textbf{RQ.1} Does the identified causal cues, namely, sentiment and aggression, enhance the generalization performance?
	\item \textbf{RQ.2} What is the importance of each causal cue in improving the generalization performance (ablation study)?
	\item \textbf{RQ.3} Which features does the \textbf{PEACE} utilize in input and whether these features are causal when compared to the other baselines? 
 % \raha{can we more focus on case study to verify the features are causal rather than focusing on interpretability}
\end{itemize}

\begin{table*}
\centering
\small
\begin{tabular}{m{2cm}m{4.5cm}m{1.7cm}m{1.7cm}m{1.8cm}m{0.0mm}} 
\toprule
\textbf{Datasets} & \textbf{Description} & \begin{tabular}[c]{@{}c@{}}\textbf{{\scriptsize Number of} }\\\textbf{ {\scriptsize Posts/Comments}}\end{tabular} & \begin{tabular}[c]{@{}c@{}}\textbf{{\scriptsize Hateful }}\\\textbf{ {\scriptsize Posts/Comments}}\end{tabular} & \begin{tabular}[c]{@{}c@{}}\textbf{{\scriptsize Percent of Hateful} }\\\textbf{ {\scriptsize Posts/Comments}}\end{tabular} &\\ 
% \hline \hline
\toprule
GAB~\cite{kennedy2018gab} & A collection of posts from the GAB social media platform & \centering 31,640 & \centering 7,657 & \centering 24.2 &\\
\hline
Reddit~\cite{qian2019benchmark} & Conversation threads from the Reddit platform & \centering 13,633 & \centering 4,219 & \centering 31 &\\
\hline
Wikipedia~\cite{wulczyn2017ex} & A collection of comments on Wikipedia website & \centering 1,13,728 & \centering 22,796 & \centering 20 &\\
\hline
Twi-Red-You & Social media comments from three sites, namely, Twitter, Reddit, and YouTube & \centering 86,283 & \centering 49,273 & \centering 57.2 &\\
\hline 
FRENK & Social media comments from Facebook targeting LGBT and Migrants & \centering 10,034 & \centering 3,592 & \centering 35.8 &\\
\bottomrule
\end{tabular}
\caption{Dataset statistics of the experimental datasets with corresponding platforms and percentage of hateful comments or posts.}
\label{datasets}
\end{table*}

\subsection{Datasets and Evaluation metrics}
% \raha{why did you use these datasets? commonly used for domain generalization?}
We perform binary classification of detecting hate speech on various widely used benchmark hate datasets. Since we aim to verify cross-platform generalization, for cross-platform evaluation, we use four datasets from different platforms: Wikipedia, Facebook, Reddit, GAB, and Twitter-Reddit-YouTube. All datasets are in the English language. Wikipedia dataset~\cite{wulczyn2017ex} is a collection of user comments from the Wikipedia platform consisting of binary labels denoting whether a comment is hateful. Reddit~\cite{qian2019benchmark} is a collection of conversation threads classified into hate and not hate. GAB~\cite{kennedy2018gab} is a collection of annotated posts from the GAB website. It consists of binary labels indicating whether a post is hateful or not. Finally, Twitter-Reddit-YouTube~\cite{kennedy2020constructing} is a collection of posts and comments from three platforms: Twitter, Reddit, and YouTube. It contains ten ordinal labels (sentiment, (dis)respect, insult, humiliation, inferior status, violence, dehumanization, genocide, attack/defense, hate speech), which are debiased and aggregated into a continuous hate speech severity score (hate speech score). We binarize this data such that any data with a hate speech score less than 0.5 is considered non-hateful and vice-versa. Although Twi-Red-You and Reddit both contain data from Reddit, these data do not necessarily have the same distribution. The distribution of datasets from the same platform can still defer due to variations in the timestamps, targets, locations, and demographic attributes. The FRENK dataset~\cite{ljubevsic2019frenk} contains Facebook comments in English and Slovene covering LGBTQ and Migrant targets. We only consider the English dataset. The dataset was manually annotated for different types of unacceptable discourses (e.g., violence, threat). We use the binary hate speech classes hate and not-hate. A summary of the datasets can be found in Table \ref{datasets}. For comparison with baseline methods, macro F-measure (F1) is used as an evaluation metric for validation.

\subsection{Baselines}
% \raha{have a new section Baselines}

\begin{itemize}

    \item \textbf{ImpCon (AugCon Variant)}~\cite{kim2022generalizable} - this baseline utilizes contrastive learning with data augmentation to map similar posts closer to each other in the representation space to enable better generalization. 
    % This model was originally proposed for implicit hate and had two variants namely, ImpCon and AugCon. ImpCon utilized implications of different hate posts to identify positive and negative samples for contrastive learning, and AugCon utilized data augmentation to do the same. Since implications are not available for our evaluation datasets, we consider the AugCon variant.\raha{why do we need to discuss implicit and explicit in so much details?just say we use this variant for explicit}
    \item \textbf{POS+EMO}~\cite{markov2021exploring} - this baseline proposed to use linguistic cues such as POS tags, stylometric features, and emotional cues derived by different words and the global emotion lexicon named, NRC lexicon~\cite{mohammad2013crowdsourcing} to enhance the generalizable capabilities for multilingual cross-domain hate speech detection.
    \item \textbf{HateBERT}~\cite{caselli2020hatebert} - finetune the BERT-base model using approximately 1.5 million Reddit messages published by suspended communities for promoting hateful content. It results in a shifted BERT model that has learned language variety and hate polarity (e.g., hate, abuse). We report the results of fine-tuned HateBERT for all the datasets. 
    \item \textbf{HateXplain}~\cite{mathew2021hatexplain} - fine-tuned using hate speech detection datasets from Twitter and Gab for a three-class classification task (hate, offensive, or normal). It combines human-annotated rationales and BERT to improve performance by reducing unintended bias toward target communities. For each dataset, we present the results of fine-tuned HateXplain.
    
\end{itemize}
 % \raha{the last two baselines are not generalization based approaches you need to point it out}
 Both HateBERT and HateXplain are not explicitly designed for generalizability but primarily for better hate speech detection. We include these baselines as they are state-of-the-art hate speech detection methods, and due to the generalization capabilities of large language models these baselines do possess better generalization~\cite{yin2021towards,kim2022generalizable}.
\subsection{Implementation Details}

Our framework \textbf{PEACE} is implemented using the Huggingface Transformers library\footnote{https://huggingface.co/docs/transformers}. For our sentiment and aggression modules, we used existing RoBERTa-base models that have been finetuned for the sentiment and aggression downstream tasks \cite{barbieri2020tweeteval}. Both these models are finetuned on a plethora of social media posts and have shown good performance in detecting sentiment and aggression in text. Moreover, we used a pre-trained RoBERTa-base model as our hate detector encoder blocks where $n=12$.  

The overall architecture was trained to utilize cross-entropy loss with class balancing and optimized with the Adam optimizer. The learning rate was set to the standard value of 0.00002, and the dropout rate was 0.2 for the best performance. For learning \textbf{PEACE}, we trained the framework on a 40 GB VRAM NVIDIA GeForce RTX 3090 GPU with the early-stopping strategy. 
% \acil{rephrase: "GPU with 40GB VRAM"}
\begin{table*}[ht!]
\centering
\small
\begin{tabular}{clccccc}
\toprule
\multicolumn{2}{c}{\textbf{Platforms}} & \multirow{2}{*}{\textbf{HateBERT}} & \multirow{2}{*}{\textbf{\begin{tabular}[c]{@{}c@{}}ImpCon \\ (AugCon variant)\end{tabular}}} & \multirow{2}{*}{\textbf{HateXplain}} & \multirow{2}{*}{\textbf{POS+EMO}} & \multirow{2}{*}{\textbf{PEACE}} \\
% \cline{1-2}
\multicolumn{1}{c}{\textbf{Source}} & \textbf{Target} &  &  &  &  &  \\ \toprule
\multirow{5}{*}{\textbf{Twi-Red-You}} & \textbf{GAB} & 0.58 & 0.58 & {\ul 0.60} & 0.54 & \textbf{0.63} \\ % \cline{2-7} 
 & \textbf{Reddit} & {\ul 0.71} & 0.64 & \textbf{0.74} & 0.54 & \textbf{0.74} \\ % \cline{2-7} 
 & \textbf{Wikipedia} & {\ul 0.71} & 0.70 & 0.70 & 0.60 & \textbf{0.78} \\ % \cline{2-7} 
 & \textbf{Twi-Red-You} & \textbf{0.96} & 0.94 & 0.92 & 0.87 & {\ul 0.95} \\ 
 & \textbf{FRENK} & 0.46 & 0.44 & {\ul 0.48} & 0.45 & \textbf{0.53} \\
 \hline

% \multicolumn{2}{c}{\textbf{Average Performance}} &  0.68 & 0.66 & {\ul 0.69} & 0.64 & \textbf{0.73} \\ \hline
\multirow{5}{*}{\textbf{GAB}}  & \textbf{GAB} & \textbf{0.84} & 0.65 & \textbf{0.84} & {\ul 0.76} & {\ul 0.76} \\ 
& \textbf{Reddit} & 0.69 & 0.64 & {\ul 0.70} & 0.56 & \textbf{0.71} \\ % \cline{2-7} 
 & \textbf{Wikipedia} & {\ul 0.74} & 0.64 & 0.70 & 0.49 & \textbf{0.78} \\ % \cline{2-7} 
 & \textbf{Twi-Red-You} & 0.61 & \textbf{0.71} & 0.61 & 0.59 & {\ul 0.70} \\ % \cline{2-7} 
 & \textbf{FRENK} & \textbf{0.71} & 0.57 & 0.60 & 0.59 & {\ul 0.69} \\
 \hline
 
% \multicolumn{2}{c}{\textbf{Average Performance}} & {\ul 0.72} & 0.64 & 0.69 & 0.60 & \textbf{0.73} \\ \hline
\multirow{5}{*}{\textbf{Reddit}} & \textbf{GAB} & 0.56 & 0.51 & {\ul 0.59} & 0.53 & \textbf{0.61} \\ % \cline{2-7} 
& \textbf{Reddit} & {\ul 0.88} & 0.84 & \textbf{0.89} & 0.59 & {\ul 0.88} \\ 
 & \textbf{Wikipedia} & {\ul 0.66} & 0.63 & 0.64 & 0.56 & \textbf{0.74} \\ % \cline{2-7} 
 & \textbf{Twi-Red-You} & 0.73 & 0.70 & {\ul 0.77} & 0.65 & \textbf{0.78} \\ % \cline{2-7} 
 & \textbf{FRENK} & 0.42 & 0.42 & 0.44 & {\ul 0.49} & \textbf{0.54} \\
 \hline
 
% \multicolumn{2}{c}{\textbf{Average Performance}} & 0.65 & 0.62 & {\ul 0.67} & 0.56 & \textbf{0.71} \\ \hline
\multirow{5}{*}{\textbf{Wikipedia}} & \textbf{GAB} & {\ul 0.65} & 0.63 & 0.64 & 0.56 & \textbf{0.68} \\ % \cline{2-7} 
 & \textbf{Reddit} & {\ul 0.73} & 0.71 & \textbf{0.74} & 0.58 & 0.72 \\ % \cline{2-7} 
 & \textbf{Wikipedia} & {\ul 0.95} & 0.93 & 0.86 & 0.94 & \textbf{0.97} \\ 
  & \textbf{Twi-Red-You} & 0.73 & 0.72 & {\ul 0.74} & 0.69 & \textbf{0.78} \\ % \cline{2-7} 
 & \textbf{FRENK} & 0.60 & 0.51 & {\ul 0.61} & 0.52 & \textbf{0.65} \\ 
 \hline
% \multicolumn{2}{c}{\textbf{Average Performance}} & {\ul 0.73} & 0.70 & 0.72 & 0.66 & \textbf{0.76} \\ \hline
\multirow{5}{*}{\textbf{FRENK}} & \textbf{GAB} & 0.65 & {\ul 0.67} & 0.63 & 0.58 & \textbf{0.69} \\ % \cline{2-7} 
 & \textbf{Reddit} & 0.62 & {\ul 0.66} & {\ul 0.66} & 0.55 & \textbf{0.71} \\ % \cline{2-7} 
 & \textbf{Wikipedia} & 0.67 & {\ul 0.76} & 0.73 & 0.53 & \textbf{0.81} \\ 
  & \textbf{Twi-Red-You} & {\ul 0.65} & {\ul 0.65} & 0.64 & 0.62 & \textbf{0.78} \\ % \cline{2-7} 
 & \textbf{FRENK} & {\ul 0.78} & \textbf{0.79} & 0.75 & 0.72 & {\ul 0.78} \\ 
 \hline
% \multicolumn{2}{c}{\textbf{Average Performance}} & 0.67 & {\ul 0.71} & 0.68 & 0.59 & \textbf{0.74} \\ 
% \bottomrule
\end{tabular}
\caption{cross-platform and in-dataset evaluation results for the different baseline models compared against \textbf{PEACE}. Boldfaced values denote the best performance and the underline denotes the second-best performance among different baselines.}
\label{results}
\end{table*}

\subsection{RQ.1 Performance Comparison}
\subsubsection{Cross-Platform Generalization} We compare the different baseline models with \textbf{PEACE} on five real-world datasets. To evaluate the generalization capabilities of the models for each dataset, we split the data into train and test tests. We train all the models on the training data for one platform and evaluate the test sets of all the platforms. Table \ref{results} demonstrates the performance comparison across the different test sets for the macro-F1 metric. The column \textbf{Platforms} showcases the Source platform on which the models were trained and the Target platforms used for evaluation. For each source dataset, we show the Average Performance of each model in both in-platform and cross-platform settings. 
% \raha{too messy can we move the next experiment to additional experiments?}
As a result, we have the following observations regarding the cross-platform performance w.r.t. RQ.1:
\begin{itemize}
    \item Overall, \textbf{PEACE} consistently yields the best performance across cross-platform evaluation for all the datasets while maintaining good in-platform macro F1. Comparing only the cross-platform performance, \textbf{PEACE} leads to a 5\% improvement when trained on the Twi-Red-You dataset, 3\% improvement for the GAB dataset, 6\% improvement for Reddit, 3\% improvement for the Wikipedia dataset, and 4\% improvement for FRENK dataset. 
    \item Among the four baselines, HateBERT serves as the strongest baseline in most cases, followed by HateXplain. This result is justified as both HateBERT and HateXplain are fine-tuned BERT models on large corpora of hateful content. We further fine-tune both HateBERT and HateXplain for each dataset. ImpCon performs well for some of the combinations, while for others, it cannot outperform HateBERT and HateXplain. We believe this is because the AugCon variant utilizes simple data augmentation. As a result, it might not be able to learn as good representations as the ImpCon variant that leverages the implications of hate. Furthermore, the utilization of the ImpCon variant is a challenging task in real-world scenarios, as the implications are not readily available for large datasets. 
    % \raha{make it a seperate bullet point}
    \item The linguistic feature-based baseline (POS + EMO) doesn't generalize well to these datasets. We argue this is because the posts in these datasets are highly unstructured and grammatically incorrect. Even after pre-processing the inferred POS tags and emotion words may not be reflective of the hate content. As a result, the reliance on these features hurts the generalization performance.
    \item Majority of the baselines attain improved performance when trained on the Wikipedia dataset. We argue this is because of the size of the dataset. Among the four datasets, Wikipedia is the largest dataset indicating that a model can generalize better when it's trained on large datasets.
\end{itemize}

\begin{table*}[ht!]
\centering
\small
\begin{tabular}{clccccc}
\toprule
\multicolumn{2}{c}{\textbf{targets}} & \multirow{2}{*}{\textbf{HateBERT}} & \multirow{2}{*}{\textbf{\begin{tabular}[c]{@{}c@{}}ImpCon \\ (AugCon variant)\end{tabular}}} & \multirow{2}{*}{\textbf{HateXplain}} & \multirow{2}{*}{\textbf{POS+EMO}} & \multirow{2}{*}{\textbf{PEACE}} \\
% \cline{1-2}
\multicolumn{1}{c}{\textbf{Source}} & \textbf{Target} &  &  &  &  &  \\ \toprule
\multirow{1}{*}{\textbf{Migrants}} & \textbf{LGBTQ} & {\ul 0.74} & 0.68 & 0.65 & 0.61 & \textbf{0.78}
 \\ % \cline{2-7} 
\multirow{1}{*}{\textbf{LGBTQ}} & \textbf{Migrants} & 0.66 & {\ul 0.67} & 0.64 & 0.58 & \textbf{0.72} \\
 \bottomrule
 \end{tabular}
  \caption{cross-target evaluation results for the different baseline models compared against \textbf{PEACE}. Boldfaced values denote the best performance among different baselines.}
  \label{cross_topic}
 \end{table*}
 
\subsubsection{Cross-Target Generalization}
Furthermore, we also conducted another experiment for the FRENK dataset to evaluate how the different models generalize in a cross-target setting, where the datasets belong to the same platform (i.e., have similar ways of expressing hate) but discuss different targets of hate. Along with the hate labels, the FRENK dataset also provides the targets of hate in the dataset, namely, \textit{LGBTQ} and \textit{Migrants}. Table \ref{cross_topic} demonstrates the performance comparison for the macro-F1 metric. 

We had the following observations regarding the cross-target generalization performance w.r.t. \textbf{RQ.1}:
\begin{itemize}
    \item Comparing the cross-target generalization, we observe that C-Hate leads to an average gain of 4\% improvement over the baselines. The results indicate that utilizing causal cues such as the overall sentiment and the aggression aids in learning generalizable representations and improve cross-target generalization performance.
    \item Across the different baselines HateBERT and ImpCon perform the best. The overall performance of HateBERT indicate that the large language models such as BERT when fine-tuned on a particular downstream task (fine-tuning BERT on hate content resulted in generation of HateBERT) can lead to competitive generalization capabilities. Furthermore, the ImpCon model performs well as it leverages data augmentation which results in more training data leading to better generalization.
\end{itemize}

\begin{figure}[ht!]
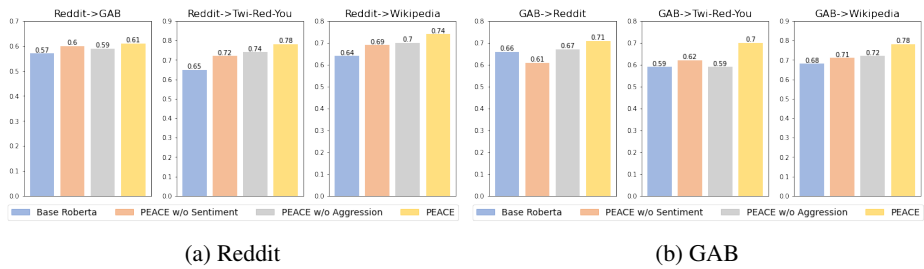
%
    \centering
    \subfloat[\centering Reddit] 
    {{\includegraphics[width=0.5\textwidth]{AblationReddit.pdf} 
    \label{ablationReddit}}%
     }%
    % \subfloat[\centering Direction]{{\includegraphics[width=0.35\textwidth]{EA_direction.pdf} 
    % \label{EAdirec}}%
    % }%
     \subfloat[\centering GAB]
     {{\includegraphics[width=0.5\textwidth]{AblationGAB.pdf} 
    \label{EAtype}}%
    }%
    \caption{Comparison of cross-platform macro-F1 score to calculate the importance of each cue compared with the final model for Reddit and GAB datasets. }
    \label{ablation}
    %\raha{legends are so small, and the two graphs are so wide apart. it does not look professional }}%
\end{figure}

% \begin{figure}[ht!]%
%     \centering
%     \quad
%     \centering
%     \subfloat[\centering Reddit] {{\includegraphics[width=0.65\textwidth]{AblationReddit.pdf} 
%     \label{ablationReddit}}%
%      }%
%     \subfloat[\centering GAB]{{\includegraphics[width=0.65\textwidth]{AblationGAB.pdf} 
%     \label{ablationGAB}}%
%     }%
   
%     \caption{Comparison of cross-platform macro-F1 score to calculate the importance of each cue compared with the final model for Reddit and GAB datasets. \raha{this figure is taking up too much space can we reorganize} }
%     \label{ablation}
%     %\raha{legends are so small, and the two graphs are so wide apart. it does not look professional }}%
% \end{figure}

\subsection{RQ.2 Importance of each cue}
To assess the individual importance of the different causal cues used in \textbf{PEACE} with regard to the performance, we conduct the following experiments. We consider three variants of \textbf{PEACE}, one which utilizes only sentiment as the causal cue, namely, \textit{Sentiment} one which utilizes only aggression as the causal cue, namely, \textit{Aggression}, and one which utilizes a RoBERTa base classifier without any causal cues, namely, \textit{Base Roberta}. We conduct cross-platform experiments by training these three variants on the Reddit and the GAB datasets. The results obtained can be seen in Figure \ref{ablation}(a) for Reddit and Figure \ref{ablation}(b) for GAB. As observed, \textbf{PEACE} performs the best when both causal cues are considered. The results can deteriorate by as little as 5\% to as high as 13\% without the inclusion of causal cues. Among the three variants, it is observed that \textbf{PEACE} mostly benefits from the aggression cue and for some datasets, it benefits from the sentiment cue. The main reason for aggression being a strong cue is because aggression and hate are very similar tasks and earlier works have shown that aggression leads to hatred~\cite{sengupta2022does}.
% Furthermore, the sentiment may sometimes be misleading \raha{I'm not comfortable with this reason, you are just contradicting your own proposed method}. This is because, when a person expresses hate towards others, the overall sentiment is always negative, whereas when the content is not hateful the sentiment may or may not be negative. For instance, the sentence "I will kill myself" possesses negative sentiment but it is not considered hateful as it is not directed towards someone else.  
However, the base model consistently does worst, indicating that the utilization of causal cues is important to enhance the generalizability performance for hate speech detection.

\begin{table*}[h]
\centering
\small
\resizebox{\columnwidth}{!}{\begin{tabular}{llm{2.2in}m{2.5in}} 
\toprule
\multicolumn{2}{l}{\multirow{2}{*}{\textbf{Model}}} & \multicolumn{2}{c}{\textbf{Platform}}                              \\ 
\cline{3-4}
\multicolumn{2}{l}{}                       & \textbf{Gab}                       & \textbf{Reddit}                        \\ 
\hline
% \hline
\multicolumn{2}{l}{\textbf{HateXplain}}                 & \includegraphics[width=2.2in]{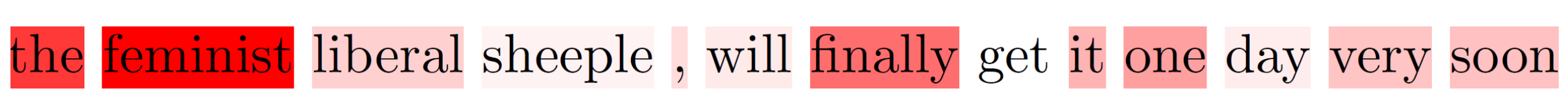} & \includegraphics[width=2.5in]{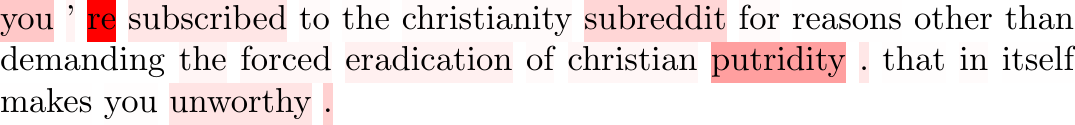}  \\ 
\hline
\multicolumn{2}{l}{\textbf{ImpCon}}                 & \includegraphics[width=2.2in]{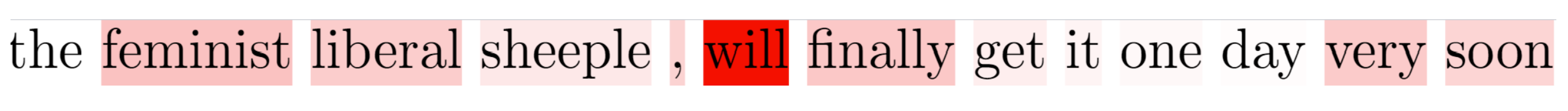} & \includegraphics[width=2.5in]{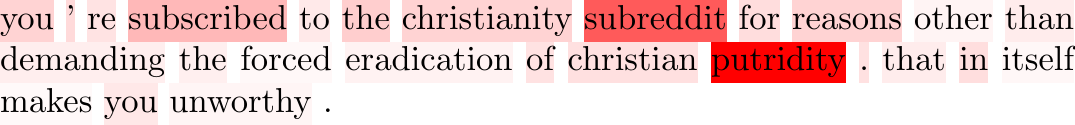}  \\ 

\hline
 & Sentiment               & \includegraphics[width=2.2in]{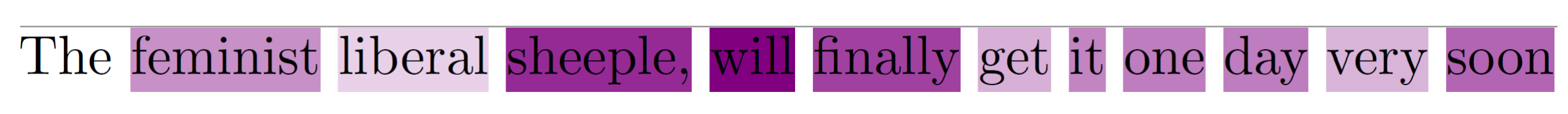}                          & \includegraphics[width=2.5in]{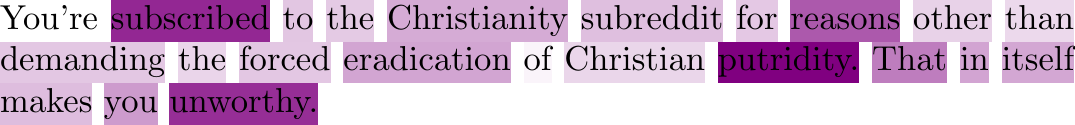}                              \\
                      &   \multicolumn{1}{c}{$+$}  & \multicolumn{1}{c}{$+$}     & \multicolumn{1}{c}{$+$}          \\
                    \textbf{Ours}  & Aggression              & \includegraphics[width=2.2in]{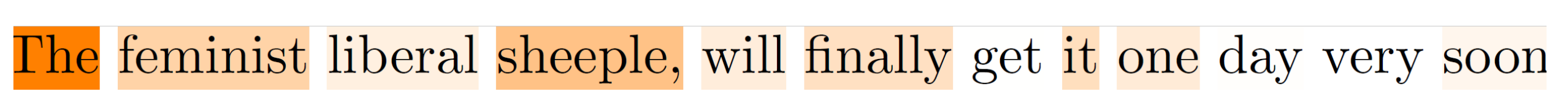}     & \includegraphics[width=2.5in]{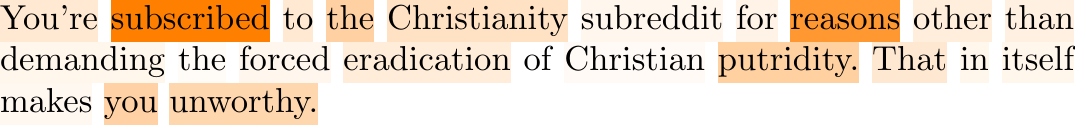}          \\
                      &  \multicolumn{1}{c}{$\downarrow$}   & \multicolumn{1}{c}{$\downarrow$}     & \multicolumn{1}{c}{$\downarrow$}          \\
                      & Full Model              & \includegraphics[width=2.2in]{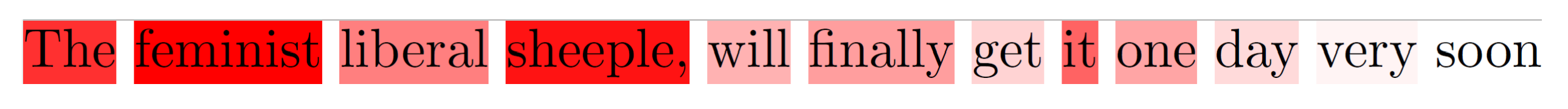}     & \includegraphics[width=2.5in]{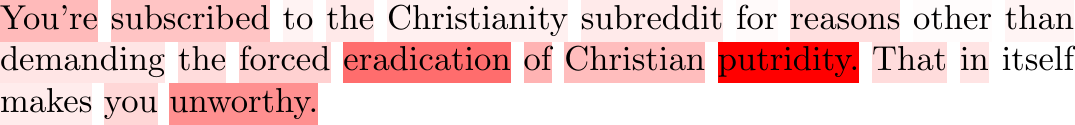}          \\
\bottomrule
\end{tabular}}
\caption{Case study illustrating the different features/tokens chosen as important tokens to detect hateful content across different models. Darker shades of the color represents the importance level of the token.}
\label{casestudy}
\end{table*}

% \acil{Rephrase the caption for table 4 -- a more descriptive caption that talks about the illustrated/color coded samples across different models would be apt here.}

\subsection{RQ.3 Case Study}

Here we provide a case study that verifies the importance of causal cues in identifying the correct context for detecting hate speech 
% \raha{you verify that it indeed focuses on the causal cues}. 
Moreover, here we visually compare \textbf{PEACE} token level attention with the baseline models HateXplain and ImpCon. In order to visualize the token importance of a given model towards its prediction, we followed a similar procedure as the cue extractor \cite{clark2019does}, where the final encoder block's attention layer was utilized to accumulate the token importance by visualizing the attention weights. 

We randomly sampled hate speech text from Reddit and Gab platforms to select candidate examples for the case study. Table \ref{casestudy} shows a few such samples with the attention token importance visualization. In the \textbf{C-Hate's} row, we annotate the sentiment module attention in  \colorbox{violet}{violet} and aggression module attention \colorbox{orange}{orange}.
The example from the Gab platform is an instance of hate towards feminist liberals. The word \textit{"sheeple"} and phrase \textit{"get it one day"}  can be considered as the deciding components of the text being hate speech. In contrast to the HateXplain and ImpCon,  \textbf{PEACE} is attending to the word "sheeple" correctly. And we see that both the sentiment and aggression modules are giving high importance to the \textit{"sheeple."} We have a similar observation about the phrase \textit{"get it one day"} where \textbf{PEACE} is successful in giving more attention to that phrase towards hate speech detection. A notable observation here is that the sentiment module is attending to the above phrase well, which could be the reason behind \textbf{C-Hate's} successfully identifying the correct context towards hate.  

The next example from the Reddit platform was a complex sentence for hate speech detection, given that hate is implied, not directly expressed. As we can see, both ImpCon and HateXplain models tend to the word \textit{"putridity"} but not to the critical contextual components that signify implicit hate, such as \textit{"forced eradication"} and \textit{"unworthy."} This example illustrates the issue in vocabulary-based approaches to generalized hate speech detection. On the contrary, we can see that the sentiment and aggression modules accurately attend to the \textit{"forced eradication"} and \textit{"unworthy"} phrases navigating \textbf{PEACE} to correctly identify the hate speech context.

\section{Limitations and Error Analysis}

In this section, we conduct an error analysis to better understand our work's limitations and aid future work in cross-platform generalized hate speech detection. For this analysis, we select the FRENK dataset (Facebook) as the testing dataset, given it contains fine-grained information about the data, such as hate targets (LGBTQ vs. migrants) and hate types (offense vs. violence). We used the \textbf{PEACE} models trained on other platforms (Twitter, Gab, Reddit, and Wiki) to run the test on the FRENK dataset mentioned above. Finally, we analyze each model's misclassification rate/error rate under dimensions of hate target and hate type. 

\begin{figure}[ht!]%
    \centering
    \subfloat[\centering target] 
    {{\includegraphics[width=0.35\textwidth]{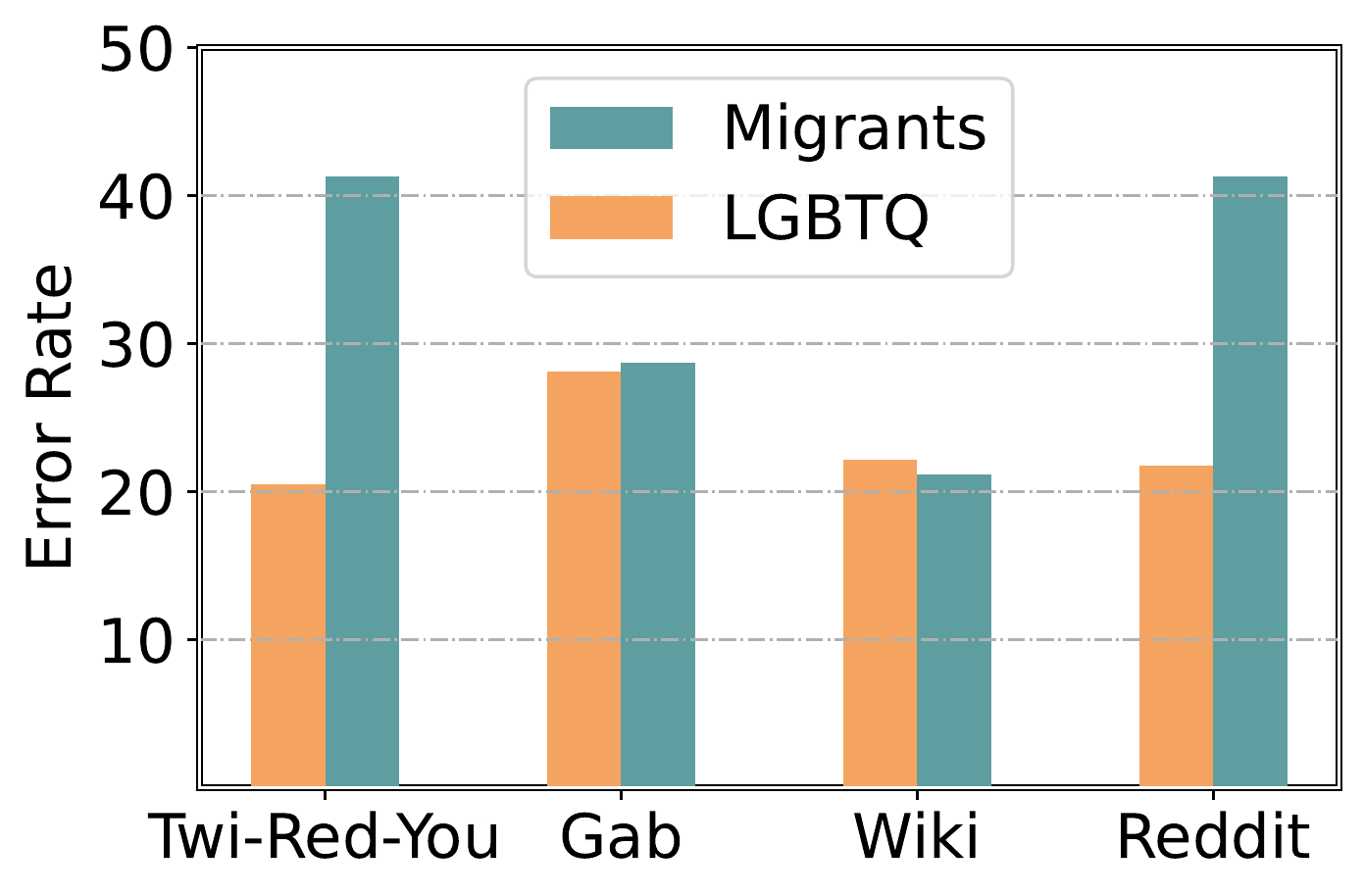} 
    \label{EAtopic}}%
     }%
    % \subfloat[\centering Direction]{{\includegraphics[width=0.35\textwidth]{EA_direction.pdf} 
    % \label{EAdirec}}%
    % }%
     \subfloat[\centering Type]
     {{\includegraphics[width=0.35\textwidth]{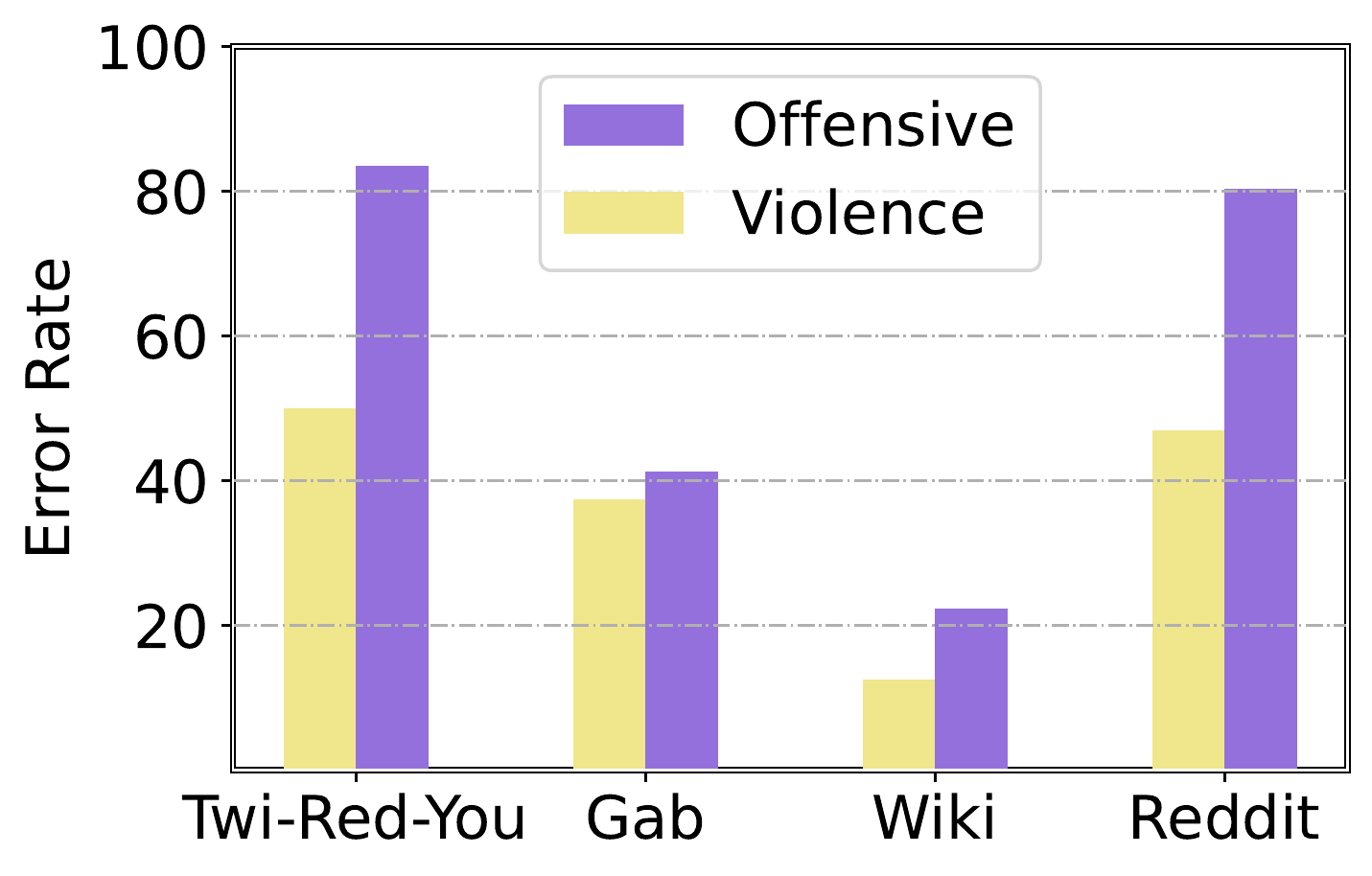} 
    \label{EAtype}}%
    }%
    \caption{Analysing error rate of \textbf{PEACE} under different Dimensions such as (a) hate targets (LGBTQ vs. migrants) and (b) hate type (offense vs. violence).}
    \label{EA}
    %\raha{legends are so small, and the two graphs are so wide apart. it does not look professional }}%
\end{figure}

\begin{table*}[ht!]
\small
\centering
\begin{tabular}{m{2cm}|m{8cm}}
\toprule
Hate Type                  & Examples \\ \hline
\multirow{3}{*}{Violence}  &   shoot them all, done!!! let the communists solve the problem!!! coz i believe that these people wont stop, sooner or later, Germany will have to use guns       \\ 
                           &   Quick... Bomb it.       \\
                           &    Send troops to reinforce the entry's in Europe, the countries they are in is in safe zones, if they continue to move forward shoot to kill, as this is regarded as invasion.      \\ \hline
\multirow{3}{*}{Offensive} &   The annoying thing is that 75\% of the migrants are Young men, why aren't they fighting for THEIR country? Or is it more a case of they can get more from European countries (money, house,education etc)       \\
                           &    Are there terrorists hidden in migration groups? Likely.
      \\
                            &   And they breed like grasshoppers.. Bye bye Europe.      \\
\bottomrule
\end{tabular}
\caption{Examples representing the different kinds of hate. The violence hate type is more explicit and direct whereas the offense hate type is more subtle and implicit.}
\label{examples}
\end{table*}

% \acil{table 5 is missing a caption}

As seen in Figure \ref{EA}(a), the model tends to have a higher error rate in detecting migrants-related samples, particularly when trained on Reddit and Twi-Red-You datasets. One notable characteristic we observed in the  Reddit and Twi-Red-You datasets is that the hate examples tend to include a majority of targeted hate towards particular individuals. Similarly, the LGBTQ target in FRENK dataset contains a majority of hate examples towards individuals. However, in contrast, the migrant target contains more generic hate examples towards a group of people. This mismatch in training and testing platforms might be causing the high error rate in the migrants compared to the LGBTQ.

According to the error analysis conducted by Figure \ref{EA}(b) we see that \textbf{PEACE} model has a higher error rate in the offensive hate type than the violence type. We further analyze this matter by examining the traits in the text that correspond to each of these hate types. 
Table \ref{examples} contains some representative samples from each of these two categories. In the violence hate type, the hate aspect is quite explicit to the reader/model. Moreover, here the sentiment and aggression cues are easily detectable. However, in the offensive hate type, we see hate to be inherently more implicit than explicit. Moreover, learning valuable signals through sentiment or aggression becomes problematic when the expressed hatred is implicit. 

% \begin{figure}[t]
%      \centering
%      \begin{subfigure}{0.3\textwidth}
%          \centering
%          \includegraphics[width=\textwidth]{EA_topic.pdf}
%          \caption{}
%          \label{fig:EAtopic}
%      \end{subfigure}
%      \hfill
%      \begin{subfigure}{0.3\textwidth}
%          \centering
%          \includegraphics[width=\textwidth]{EA_direction.pdf}
%          \caption{}
%          \label{fig:EAdirec}
%      \end{subfigure}
%      \hfill
%      \begin{subfigure}{0.3\textwidth}
%          \centering
%          \includegraphics[width=\textwidth]{EA_type.pdf}
%          \caption{}
%          \label{fig:EAtype}
%      \end{subfigure}
%         \caption{}
%         \label{fig:erroranalysis}
% \end{figure}

\section{Conclusions and Future Work}
% \HL{Remember to emphasize this causality-guided framework as our primary contribution.} 

The widespread popularity and easy accessibility of online social media platforms have led humans to easily share their opinions with the rest of the world. However, some people misuse this privilege to spread hateful content targeted to denigrate an individual or group. As a result, automated hate speech detection has become a crucial task. However, due to various factors, such as the evolving nature of hate and the limited availability of labeled data in a platform, it is challenging to develop a generalizable hate speech detection model.
% However, current state-of-the-art models exhibit poor generalization capabilities failing to generalize to new topics and different ways of expressing hate. 
To address the poor generalization problem, in this paper, we proposed a generalizable hate speech detection model, named \textbf{PEACE}, that considers the inherent causal cues that characterize whether a text content is hateful. Studies in various disciplines, such as sociology and psychology, indicate that hateful content contains specific inherent cues that can be leveraged and quantified better to detect hate speech across cross-platform and cross-target settings. We leverage the text's aggression and the content's overall sentiment to learn generalizable representations for improved hate speech detection. We conducted extensive experiments and showed that \textbf{PEACE} can generalize better across five different social media platforms and two different targets when compared with various state-of-the-art baselines. We further conducted experiments to show the importance of each causal cue and case study to identify the features \textbf{PEACE} relies on for detecting hate speech.

\textbf{PEACE}'s generalization prowess comes from the two primary causal cues, which are manually identified. One potential direction would be to investigate how to automate identifying the cues and build an end-to-end system. Moreover, hate speech detection can be further enriched by considering the context of the conversation. Another direction would be to explore how to leverage context in a cross-platform setting to improve generalization capabilities further.

\section{Ethical Statement}

\subsubsection{Freedom of Speech and Censorship}
 
Our research on cross-platform hate speech detection aims to develop algorithms that can effectively identify and mitigate harmful language across multiple platforms. We recognize the importance of protecting individuals from the adverse effects of hate speech and the need to balance this with upholding free speech. Content moderation is one application where our method could detect and censor hate speech on social media platforms such as Twitter, Facebook, Reddit, etc. However, one ethical concern is our system's false positives, i.e., if the system incorrectly flags a user's text as hate speech, it may censor legitimate free speech. Therefore, we discourage incorporating our methodology in a purely automated manner for any real-world content moderation system until and unless a human annotator works alongside the system to determine the final decision. 

\subsubsection{Use of Hate Speech Datasets}

In our work, we incorporated publicly available well-established datasets. And we have correctly cited the corresponding dataset papers and followed the necessary steps in utilizing those datasets in our work. Moreover, we understand that the hate speech examples used in the paper are potentially harmful content that could be used for malicious activities. However, our work aims to help better investigate, comprehend, and help mitigate the harms of online hate. Therefore, we have assessed that the benefits of incorporating these real-world examples to explain our work better outweigh the potential risks.

\subsubsection{Fairness and Bias in Detection}

Our work strives to prioritize using natural language processing tools for social good while respecting the principles of fairness and impartiality. To reduce biases and ethical problems, we openly disclose our methodology, results, and limitations and will continue to assess and improve our system in the future.

\section{Acknowledgements}
This material is based upon work supported by, or in part by the Office of Naval Research (ONR) under
contract/grant number N00014-21-1-4002 and the Army Research Office under the grant number W911NF2110030.

\bibliographystyle{splncs04}
\bibliography{mybibliography}
\end{document}